\documentclass[lettersize,journal]{IEEEtran}

\usepackage{orcidlink} 

\usepackage{placeins}
\usepackage{microtype}
\usepackage{multicol}
\usepackage{url}
\usepackage{hyperref}
\usepackage{comment}
\usepackage{lastpage}

\usepackage{amsmath}
\usepackage{amsthm}
\usepackage{amssymb}
\usepackage{amsopn}
\usepackage{bbm}
\usepackage{bm}
\usepackage{nicefrac}
\usepackage{mathtools}

\usepackage{graphicx}
\usepackage{float}
\usepackage{subcaption} % Note: loads caption
\usepackage{algorithm}
\usepackage{algorithmicx}
\usepackage{algpseudocode}

\theoremstyle{plain}
\newtheorem{theorem}{Theorem}[section]
\DeclareMathOperator{\Ii}{\mathcal{I}}

\DeclareMathOperator{\RR}{\mathbb{R}}
\DeclareMathOperator{\Sp}{\mathbb{S}}

\DeclareMathOperator{\aaa}{\mathbf{a}}
\DeclareMathOperator{\bb}{\mathbf{b}}

\DeclareMathOperator{\ff}{\mathbf{f}}
\DeclareMathOperator{\gb}{\mathbf{g}}
\DeclareMathOperator{\hh}{\mathbf{h}}

\DeclareMathOperator{\uu}{\mathbf{u}}
\DeclareMathOperator{\vvv}{\mathbf{v}}
\DeclareMathOperator{\ww}{\mathbf{w}}
\DeclareMathOperator{\xx}{\mathbf{x}}
\DeclareMathOperator{\yy}{\mathbf{y}}

\DeclareMathOperator{\Xf}{\mathfrak{X}}
\DeclareMathOperator{\Yf}{\mathfrak{Y}}

\algnewcommand\Define{\item[{\textbf{Define:}}]}

\graphicspath{{./figures/}}

\begin{document}

\title{Sliding Window Informative Canonical Correlation Analysis}

\author{Arvind Prasadan\,\orcidlink{0000-0002-2521-7113}%
\thanks{Arvind Prasadan is with Sandia National Laboratories, 7011 East Avenue, Livermore, CA, USA 94550-9610 (e-mail: \href{mailto:aprasad@sandia.gov}{aprasad@sandia.gov})}%
%\thanks{Manuscript received MMMM DD 2025; revised MMMM DD YYYY.}%
}

%\markboth{IEEE Transactions on Signal Processing, Vol. 00, 0000}%
%{{Prasadan}: Sliding Window Informative Canonical Correlation Analysis}


\maketitle

\begin{abstract}%   <- trailing '%' for backward compatibility of .sty file
Canonical correlation analysis (CCA) is a technique for finding correlated sets of features between two datasets. 
Formally, CCA finds maximally correlated linear combinations of features in each dataset. 
In this paper, we propose a novel extension of CCA to the online, streaming data setting: Sliding Window Informative Canonical Correlation Analysis (SWICCA). 
Our method uses a streaming principal component analysis (PCA) algorithm as a backend and uses these outputs combined with a small sliding window of samples to estimate the CCA components in real time. 
The SWICCA method is applicable and scalable to extremely high dimensions, high data rates, and performs well under data drift.
Moreover, unlike prior streaming CCA algorithms, our method allows for ranks greater than one, that is, the estimation of more than one correlated set of features. 
We motivate and describe our algorithm, provide numerical simulations to characterize its performance, and provide a theoretical performance guarantee. 
We provide a real-data example on a multi-camera video dataset that demonstrates both scaling and adaptation to drift. 
\end{abstract}

%\begin{keyword}[class=MSC]
%\kwd[Primary ]{62H20}
%\kwd{62H25}
%\kwd[; secondary ]{62J10}
%\kwd{62L10}
%\end{keyword}


\section{Introduction}

Given two datasets, canonical correlation analysis (CCA)  is a general technique for finding the linear combinations of features in both datasets that are maximally correlated and can be thought of as an analogue of principal component analysis (PCA) for the cross covariance matrices from two different sets of features \cite{knapp1978canonical}. 
CCA has a long history in classical statistics \cite{hotelling1936relations} and has been used in several application domains, including signal processing \cite{ge2009does}, finance \cite{todros2012measure}, machine learning \cite{dhillon2011multi}, psychology \cite{knapp1978canonical}, and cybersecurity \cite{jadidi2023correlation}. 

The performance and convergence of CCA has been analyzed in the static, fully observed data setting, e.g., including in \cite{pezeshki2004empirical, ge2009does, bao2019canonical}, with more recent work focusing on the simultaneously high dimensional and low sample regime.
There have also been several extensions to CCA, including a sparse CCA algorithm proposed in \cite{mai2019iterative} and kernel CCA methods in \cite{akaho2006kernel, fukumizu2007statistical}. 
Recent work has studied CCA in a high dimensional setting with a two-stage algorithm \cite{asendorf2017improved}, somewhat similar in spirit to the two-stage sparse PCA algorithm presented in \cite{paul2012augmented}. 

In this work, we present a novel CCA algorithm intended for the online, streaming data setting. 
There are existing approaches to streaming or stochastic CCA (see \cite{arora2017stochastic, bhatia2018gen, gao2019stochastic, meng2021online}) that we seek to improve upon in the following ways. 
First, our goal is to develop an algorithm that is adaptive to changes in the underlying data distributions as opposed to a method that seeks to solve the static CCA problem in a memory-constrained, streaming setting. 
If there is a distributional shift in the data stream, aggregating samples across this shift as a static method would do is nonsensical. 
Second, we seek a method that processes each sample only once, that is, previous samples are not retained forever and multiple passes over the dataset is not an option.
Indeed, in a setting with data drift, after a certain point, older samples are no longer representative of the current data distribution, and in a setting with high data throughput, memory constraints might preclude storing the entire sample path. 
Third, we seek to solve the CCA problem directly, as opposed to solving an approximation or convex relaxation of the CCA objective.
Fourth, unlike several existing stochastic methods that natively track only a single canonical component, we require a method that can extract multiple correlated components simultaneously without computationally expensive deflation steps.
Hence, we present a method that is fully compatible with strictly bounded storage environments, adapts rapidly to distributional drift, and is specifically designed for real-time signal tracking applications. 
Our method is inspired by the work in \cite{asendorf2017improved}, and uses a two-step procedure where a streaming PCA method is used to preprocess the data stream before CCA is performed on the output. 

We compare our method to the state-of-the-art Gen-Oja method from \cite{bhatia2018gen} and demonstrate that while slightly more computationally demanding, we are able to do much better in terms of empirical performance under a broad range of simulated conditions. 
While Gen-Oja is not explicitly designed for the CCA problem and has utility far beyond CCA, it is directly comparable to our method, as Gen-Oja solves a generalized eigenvalue problem in the stream. 
Moreover, unlike competing prior methods, we demonstrate that our method is extremely scalable to high dimensions and high data throughput rates.

This paper is organized as follows. 
In section \ref{sec:model}, we introduce the data model and problem statement, as well as the motivating derivations needed for our proposed method. 
We additionally describe a modification to the static CCA solution and the ICCA algorithm from \cite{asendorf2017improved} that enables scaling and computation in extremely high dimensions. 
In section \ref{sec:SWICCA}, we introduce our sliding window informative CCA (SWICCA) algorithm, detailed in Algorithm \ref{alg:swicca}.
In section \ref{sec:theory}, we provide theoretical characterizations of the SWICCA algorithm, including an analysis of the computational complexity and of the error in the output; proofs of our results are deferred to the appendix. 
In section \ref{sec:experiments}, we provide simulation studies to numerically validate our method, including two applications to real data. 
Finally, in section \ref{sec:conclusions}, we provide some concluding thoughts and discuss future directions of this work. 

\section{Data Model and Problem Statement}\label{sec:model}

We operate in the streaming setting where we seek to minimize our storage of past samples. 
At time $t$, we observe a pair of samples $(\xx_t, \yy_t)$ drawn from the underlying $p$- and $q$-dimensional random vectors $\Xf_t$ and $\Yf_t$, respectively. 
That is, there is a one-to-one correspondence between samples $\xx_t$ and $\yy_t$, and without loss of generality, we assume that this `alignment' has been done and that $\Xf_t$ and $\Yf_t$ have zero mean. 
In the streaming setting, we allow for the distributions of $\Xf_t$ and $\Yf_t$ to change over time; in what follows, we suppress the time index $t$ where possible to allow for notational clarity. 

The canonical correlation analysis problem looks for directions $\ff_k \in \Sp^{p - 1}$ and $\gb_k \in \Sp^{q - 1}$, where $\Sp^{p - 1}$ denotes the unit sphere in $\RR^p$, such that $\ff_k$ and $\gb_k$ maximize the correlation between linear combinations of the random vectors, given by
\begin{equation}
\textrm{corr}\left(\ff_k^\top \Xf, \gb_k^\top \Yf\right)
\end{equation}
More than one pair of directions can be obtained by constraining subsequent pairs to be uncorrelated with previous directions. Let $\Sigma_x \in \RR^{p \times p}$ and $\Sigma_y \in \RR^{q \times q}$ be covariances of $\Xf$ and $\Yf$ respectively and let $\Sigma_{xy}$ be the cross-covariance matrix. Then, we have that the $\ff$ are eigenvectors of 
\begin{subequations}
\begin{equation}
\Sigma_x^{-1} \Sigma_{xy} \Sigma_y^{-1} \Sigma_{yx},
\end{equation}
and that the $\gb$ are eigenvectors of
\begin{equation}
\Sigma_y^{-1} \Sigma_{yx} \Sigma_x^{-1} \Sigma_{xy}.
\end{equation}
\end{subequations}
Hence, our task is to estimate and update the directions $\ff_k$ and $\gb_k$ in real-time as samples $(\xx_t, \yy_t)$ arrive, while minimizing storage needs and computational complexity. 

\subsection{CCA via the Singular Value Decomposition} 

Before proceeding to the streaming setting, we engage in a hypothetical detour that will be fruitful later. 
Suppose that we were not in the streaming setting, but instead observed all of the data, say, $n$ sample pairs. 
If we collected these into matrices $X \in \RR^{n \times p}$ and $Y \in \RR^{n \times q}$ (where the samples are rows), we would be able to write the singular value decomposition (SVD) of $X = U_x S_x V_x^\top$ and $Y = U_y S_y V_y^\top$. If we defined 
\begin{equation} \label{eq:CCA_C}
C = V_x U_x^\top U_y V_y^\top,
\end{equation}
and let $C$ have an SVD $C = W L H^\top$, we would have that the CCA directions would be given by 
\begin{equation}
\ff_k \propto \Sigma_x^{-1/2} \ww_k \textrm{ and } \gb_k \propto \Sigma_y^{-1/2} \hh_k,
\end{equation}
where $\ww_k$ and $\hh_k$ are the $k^{th}$ left and right singular vectors of $C$, respectively. 
We also have that 
\begin{equation}
\Sigma_{x}^{-1/2} = V_x S_x^{-1/2} V_x^\top \textrm{ and } \Sigma_{y}^{-1/2} = V_y S_y^{-1/2} V_y^\top.
\end{equation}

\subsection{Low-Rank Data and Informative CCA}

We assume that the $\xx_t$ and $\yy_t$ each come from a low rank subspace, say with dimensions $r_x$ and $r_y$, respectively. 
That is, if the data were fully observed, we would be able to write
\begin{equation}
X = \sum_{k = 1}^{r_x} \sigma_{x, k} \uu_{x, k} \vvv_{x, k}^\top \textrm{ and } Y = \sum_{k = 1}^{r_y} \sigma_{y, k} \uu_{y, k} \vvv_{y, k}^\top,
\end{equation}
where $\uu_{x, k}$ is the $k^{th}$ column of the matrix $U_x$ (and so on) and the $\sigma_{x, k}$ and $\sigma_{y, k}$ are the decreasing, non-negative sequence of singular values contained in $S_x$ and $S_y$ respectively. 

In this setting, the innovation of \cite{asendorf2015improved,asendorf2017improved} was to realize that when $X$ and $Y$ are low-rank and corrupted by noise, better performance can be obtained by replacing $X$ and $Y$ with their low-rank approximations, thereby obtaining the ICCA (Informative CCA) algorithm, which notably provides improved estimates of the underlying canonical vectors in these noisy regimes.
That is, in the computation of the CCA matrix $C$ (defined in Equation \ref{eq:CCA_C}), replacing the four matrices with their trimmed versions (the first $r_x$ and $r_y$ columns) leads to better performance in the presence of noise, especially in a high-dimensional setting. 
Indeed, given knowledge that the data are low-rank, this trimming is a natural step to take, even without the presence of noise. 

\subsubsection{Scaling ICCA to high dimensional settings}

One challenge in the high-dimensional setting (when $p$ and $q$ are large) is that the CCA matrix $C$ (defined in Equation \ref{eq:CCA_C}) is extremely large and dense; forming $C$ let alone directly computing its SVD can be infeasible, an observation and workaround first detailed by \cite{nadakuditi2010fundamental, nadakuditi2011fundamental}.
We can usually compute the first few singular values and vectors of $X$ and $Y$ without too much difficulty, even in the high dimensional setting, assuming that we can store $X$ and $Y$.
We then see that the matrix $U_x^\top U_y$ is an $r_x \times r_y$ matrix. 
If we write the SVD of $U_x^\top U_y$ as $A D B^\top$, we see that 
\begin{equation}
C = \left(V_x A\right) D \left(V_y B\right)^\top,
\end{equation}
where we note that since $V_x$ and $A$ have orthonormal columns, so must $V_x A$: 
\begin{equation}
(V_x A)^\top (V_x A) = A^\top V_x^\top V_x A = A^\top A = \Ii_{r_x}.
\end{equation}
A similar conclusion holds for $V_y B$. 
It then follows that the above is in fact the (truncated) SVD of $C$, so that we may immediately conclude that 
\begin{subequations}\label{eq:avoid_C}
\begin{equation}
\ff_k \propto V_x S_x^{-1/2} V_x^\top V_x \aaa_k = V_x S_x^{-1/2} \aaa_k,
\end{equation}
and
\begin{equation}
\gb_k \propto V_y S_y^{-1/2} V_y^\top V_y \bb_k = V_y S_y^{-1/2} \bb_k,
\end{equation}
\end{subequations}
where we have used the trimmed versions of the SVD of $X$ and $Y$, and $\aaa_k$ and $\bb_k$ denote the $k^{th}$ columns of $A$ and $B$, respectively. 
Hence, we may easily scale ICCA (and CCA, where we would have an $n \times n$ matrix for $U_x^\top U_y$ instead of an $r_x \times r_y$ matrix) to high dimensional settings by not forming $C$ and by taking the SVD of $U_x^\top U_y$.

\section{Sliding Window Informative CCA} \label{sec:SWICCA}

So far, we have described CCA in the static setting and have described an innovation that improves performance in the static, low-rank setting. 
However, we are really interested in the streaming setting, that is, incrementally updating the CCA directions $\ff$ and $\gb$ as each new sample $\xx_t$ comes in. 
Note that in (\ref{eq:CCA_C}), $V_x$ and $V_y$ are matrices containing the principal components: there is ample work on streaming principal components analysis (PCA) and subspace tracking that provides estimates of these quantities in the stream (see \cite{balzano2018streaming} for examples). 
However, we do not have access to $U_x$, $U_y$, $S_x$, and $S_y$. 

Nonetheless, there is hope: we do not need to know all of $U_x$ and $U_y$, but rather a matrix of their inner products. 
That is, we care about what is essentially a correlation matrix of the transformed and scaled coordinates from each dataset. 
If we assume that the data are sampled uniformly at random, that is, that the sequences $\xx_t$ and $\yy_t$ are comprised of independent and identically distributed elements and that the ordering does not matter, then any window or subsampling of the data should `look' like any other window of the same data. 
Hence, up to some scaling that depends on the window size, if we stored a window of size $w$ of the samples, $X_w \in \RR^{w \times p}$ and $Y_w \in \RR^{w \times q}$, the columns of $X_w V_x$ and $Y_w V_y$ are proportional to the columns in the corresponding sub-matrix of $U_x$ and $U_y$ respectively. 
Moreover, the norms of these columns are proportional to the singular values $\sigma_{x, k}$ and $\sigma_{y, k}$, where the scaling factor does not depend on $k$. 
If there is drift in the datasets, our approach is still reasonable, as a streaming PCA method would track the changes in the underlying data distributions, and for a reasonable window size, the matrix of loadings would still be sensible. 

Hence, we present the sliding window informative CCA (SWICCA) algorithm in Algorithm \ref{alg:swicca}.
We note that if there is drift in the data or if the data dimension is very large, it may be advantageous to store a window of loadings $\xx_t^\top \widehat{V}_x$ (and similarly for $\yy_t$) instead of computing the loadings for the window in each iteration. 
Additionally, if the data streams do not have zero mean, as part of the PCA updates we might keep track of the means, e.g., by an exponentially weighted moving average or by computing the sample mean on the window. 

\begin{algorithm}
\caption{The Sliding Window Informative CCA Algorithm}\label{alg:swicca}
\begin{algorithmic}[1]
\Define Dimensions $p = \textrm{dim}(X)$ and $q = \textrm{dim}(Y)$
\Require Rank $r_x$ such that $1 \leq r_x \leq p$
\Require Rank $r_y$ such that $1 \leq r_y \leq q$
\Require Window size $w$ such that $\max\{r_x, r_y\} \leq w$
\State Initialize streaming PCA algorithms for datasets $X$ and $Y$
\State Initialize sliding window for samples

\ForAll{Samples $(\xx_t, \yy_t)$}
\State Update streaming PCA estimates $\widehat{V}_x \in \RR^{p \times r_x}$ and $\widehat{V}_y \in \RR^{q \times r_y}$
\State Add $(\xx_t, \yy_t)$ to the window; drop the oldest item if current window size exceeds $w$

\State Let $X_w$ denote the matrix with $p$ columns and up to $w$ rows formed from the samples $\xx_t$ in the window and similarly for $Y_w$
\State Let $\widehat{U}_x \in \RR^{w \times r_x}$ be comprised of the normalized (unit $\ell_2$ norm) columns of $X_w \widehat{V}_x$
\State Let $\widehat{S}_x \in \RR^{r_x \times r_x}$ be the diagonal matrix whose non-zero entries are the $\ell_2$ norms of the columns of $X_w \widehat{V}_x$
\State Form $\widehat{U}_y$ and $\widehat{S}_y$ similarly

\State Form $\widehat{U}_x^\top \widehat{U}_y$ and take its SVD $\widehat{A} \widehat{D} \widehat{B}^\top$
\State Compute the directions $\ff_k$ and $\gb_k$ and scale them to unit $\ell_2$ norm:
\begin{equation}
\widehat{\ff}_k = \frac{\widehat{V}_x \widehat{S}_x^{-1/2} \widehat{\aaa}_k}{\left\| \widehat{V}_x \widehat{S}_x^{-1/2} \widehat{\aaa}_k \right\|_2},\textrm{ and }
\gb_k = \frac{\widehat{V}_y \widehat{S}_y^{-1/2} \widehat{\bb}_k}{\left\| \widehat{V}_y \widehat{S}_y^{-1/2} \widehat{\bb}_k \right\|_2},
\end{equation}
\State Project the window of data onto the directions and compute the empirical correlations; alternatively, the diagonal entries of $\widehat{D}$ are estimates of the correlations
\EndFor
\end{algorithmic}
\end{algorithm}

\section{Theoretical Results}\label{sec:theory}

In this section, we provide theoretical characterizations of the performance of our method. 
We defer the proofs of our results to the Appendix, namely sections \ref{app:complexity} and \ref{app:error}. 

\subsection{Computational Complexity}

We provide a brief sketch of the computational complexity of our method, assuming the use of the PIMC and GROUSE streaming PCA methods \cite{balzano2018streaming}, both of which are amenable to drifting data distributions; other methods may have slightly different costs, but with these two methods, neither method affects the final complexity results. 
We summarize our results as follows:
\begin{theorem}\label{thm:complexity}
The SWICCA algorithm, applied to a $p$ dimensional dataset with rank $r_x$ and a $q$ dimensional dataset with rank $r_y$, with a window size $w$ and either the PIMC or GROUSE streaming PCA methods as a backend has a per-update time complexity of
\begin{subequations}
\begin{equation}
\mathcal{O}\left(w \left[p r_x + q r_y\right]\right),
\end{equation}
and a space complexity of
\begin{equation}
\mathcal{O}\left(w \left[p + q \right]\right).
\end{equation}
\end{subequations}
Moreover, if we decide to store a window of loadings, rather than of samples, the time complexity of each iteration drops to 
\begin{subequations}
\begin{equation}
\mathcal{O}\left(p r_x^2 + q r_y^2 + w \left[r_x + r_y\right] + r_x r_y \min\{r_x, r_y\}\right),
\end{equation}
and the space complexity drops to 
\begin{equation}
\mathcal{O}\left(\max\{p, w\} r_x + \max\{q, w\} r_y\right).
\end{equation}
\end{subequations}
\end{theorem}

For comparison, the Gen-Oja algorithm, strictly as written in \cite[Algorithm~1]{bhatia2018gen}, explicitly constructs the $(p + q) \times (p + q)$ block matrices $A$ and $B$.
This naive formulation requires $\mathcal{O}\left(p^2 + q^2 + p q\right)$ time to form the matrices, followed by $\mathcal{O}\left(p^2 + q^2\right)$ for the subsequent matrix-vector multiplications, resulting in an overall $\mathcal{O}\left(p^2 + q^2\right)$ computational complexity per iteration.
Similarly, the explicit storage requirements scale to $\mathcal{O}\left(p^2 + q^2\right)$ to hold these matrices in memory.
However, to ensure a rigorous and fair empirical baseline in our experiments, we implemented a heavily optimized version of Gen-Oja that does not form these large matrices.
By leveraging the associativity of matrix-vector multiplication---for example, computing rank-1 updates of the form $(x x^T)v$ as the functionally equivalent $x(x^T v)$---the explicit construction of the large covariance matrices can be entirely bypassed.
This algorithmic optimization reduces Gen-Oja's operational implementation footprint to strictly $\mathcal{O}(p + q)$ in both time and space per iteration.
While this optimized baseline allows Gen-Oja to operate with extreme efficiency, it remains mathematically restricted to extracting a single canonical component natively.
Relative to this highly optimized baseline, SWICCA provides a powerful generalization; it extracts multiple correlated components ($r > 1$) simultaneously while preserving a highly scalable $\mathcal{O}(w[p + q])$ memory footprint and remaining entirely adaptable to data drift.

\subsection{Error Analysis}\label{sec:performance}

We now provide a performance analysis of the output from the SWICCA algorithm. 
In what follows, we will show that if the streaming PCA algorithm yields accurate or consistent estimates of the principal components and if the noise level in the data is not too high (relative to the error in the PCA estimates), the SWICCA algorithm will produce accurate estimates. 

\subsubsection{Setup}\label{ssec:setup}

We assume that we have a window of $w$ observations of the random variables $\Xf$ and $\Yf$, collected into matrices $X \in \RR^{w \times p}$ and $Y \in \RR^{w \times q}$.
We assume that these observations are corrupted by noise, such that we may decompose the window of data as 
\begin{subequations}
\begin{equation}
X = U_x S_x V_x^\top + G_x \in \RR^{w \times p}
\end{equation}
and
\begin{equation}
Y = U_y S_y V_y^\top + G_y \in \RR^{w \times q},
\end{equation}
\end{subequations}
where $V_x \in \RR^{p \times r_x}$ and $V_y \in \RR^{p \times r_y}$ are the underlying principal components and $G_x$ and $G_y$ are matrices of noise.
Hence, $U_x S_x V_x^\top$ and $U_y S_y V_y^\top$ are the true underlying signals. 

For our analysis, we make the following assumptions. 
We assume that we know the ranks $r_x$ and $r_y$ and that the window size $w$ is fixed and is greater than the ranks. Moreover, we assume that the singular values in $S_x$ and $S_y$ are strictly lower bounded by some constant $c_{\sigma} > 0$ and upper bounded by a constant $C_{\sigma} > c_{\sigma}$. 
We also require that there are $1 \leq r_C \leq \min\{r_x, r_y\}$ correlated pairs of directions, that the absolute values of the correlations are lower bounded by some constant $c_{\rho} > 0$, and that the number of correlated components, like the ranks, is fixed. 

We allow the singular values and correlations to drift, as long as their magnitudes are lower bounded. 
The principal components of the dataset may also drift. 

\subsubsection{Theorem statement}

Given whatever streaming PCA method we choose, we obtain estimates 
\begin{equation}\label{eq:v_delta}
\widehat{V}_x = V_x + \Delta_x \textrm{ and } \widehat{V}_y = V_y + \Delta_y
\end{equation}
of the principal components at the end of the window. 
Note that the estimated matrices as well as the original matrices of principal components have orthonormal columns, and that $\Delta_x$ and $\Delta_y$ denote the error in our estimate. 
When we discuss the error of a vector, especially of a unit-norm eigenvector, we assume that the inner product between a vector $\uu$ and its estimate $\widehat{\uu}$ is positive: $\uu^\top \widehat{\uu} \geq 0$. 
Let $F = \begin{bmatrix} \ff_1 & \cdots & \ff_{r_x}\end{bmatrix} \in \RR^{p \times r_x}$ and $G = \begin{bmatrix} \gb_1 & \cdots & \gb_{r_y}\end{bmatrix} \in \RR^{q \times r_y}$ be the matrices whose columns are the true correlated canonical directions and let $L \in \RR^{r_x \times r_y}$ be the diagonal matrix containing the true correlations. 
We make the following claim:
\begin{theorem}\label{thm:error}
If we have that
\begin{equation}
\max\left\{\left\|\Delta_x\right\|_F, \left\|\Delta_y\right\|_F, \left\|X \Delta_x\right\|_F, \left\|Y \Delta_y\right\|_F\right\} \rightarrow 0,
\end{equation}
where $\Delta_x$ and $\Delta_y$ are as defined in (\ref{eq:v_delta}), $X$ and $Y$ are as in Section \ref{ssec:setup}, and the remainder of the assumptions in Section \ref{ssec:setup} hold, we have that 
\begin{equation}
\left\|\widehat{F} - F\right\|_F, \left\|\widehat{G} - G\right\|_F, \left\|\widehat{L} - L\right\|_F \rightarrow 0.
\end{equation}
\end{theorem}
Note that if we have distributional drift of the principal components or the singular values, we would need to use a streaming PCA method that can handle that drift. 
Since the principal components are updated with each incoming sample and the loadings are computed for that sample, if the error at each sample is vanishingly small, our theorem will hold.

\section{Simulation Study}\label{sec:experiments}

\subsection{Synthetic Data}\label{ssec:synth}

The primary advantage of our proposed method is its ability to extract and track multiple correlated components dynamically in non-stationary environments.
To demonstrate this, we generate two synthetic datasets. 
These initial simulations serve as controlled, deliberately low-dimensional ``toy'' examples explicitly designed to visualize subspace tracking dynamics and adaptation behaviors prior to evaluating massive real-world streams. 
The first dataset has $p=100$ covariates and an intrinsic rank of 2; the second has $q=50$ covariates and an intrinsic rank of 3.
Without loss of generality, the mean of both datasets is zero. 

The underlying canonical correlation coefficients between the first two principal components of each dataset are fixed at $\rho_1 = 0.8$ and $\rho_2 = 0.5$. 
For all settings, we generate a streaming sample path of $n = 1000$ samples and average our results over $50$ independent trials. 
In each trial, we add white Gaussian noise with standard deviation $0.1 / \sqrt{n}$ to both datasets.

We evaluate the algorithms under two primary environmental conditions:
\begin{enumerate}
\item \textbf{Static Subspace:} The underlying data distribution and component loadings remain constant over time.
\item \textbf{Continuous Drift:} We fix the correlation structure but continuously rotate the principal components of each dataset at every time step, such that the principal components at the end of the stream are orthogonal to those at the beginning. 
\end{enumerate}

We compare SWICCA against the Generalized Oja (Gen-Oja) algorithm \cite[Algorithm~1]{bhatia2018gen}.
We use the recommended step functions for Gen-Oja ($\alpha_t \propto 1 / \log(t)$ and $\beta_t \propto 1 / t$) and configure it to track the top canonical component ($r=1$). 
For SWICCA, we leverage the GROUSE streaming PCA algorithm to handle the continuous drift.

\begin{figure}[tb]
\centering
\includegraphics[width=\linewidth]{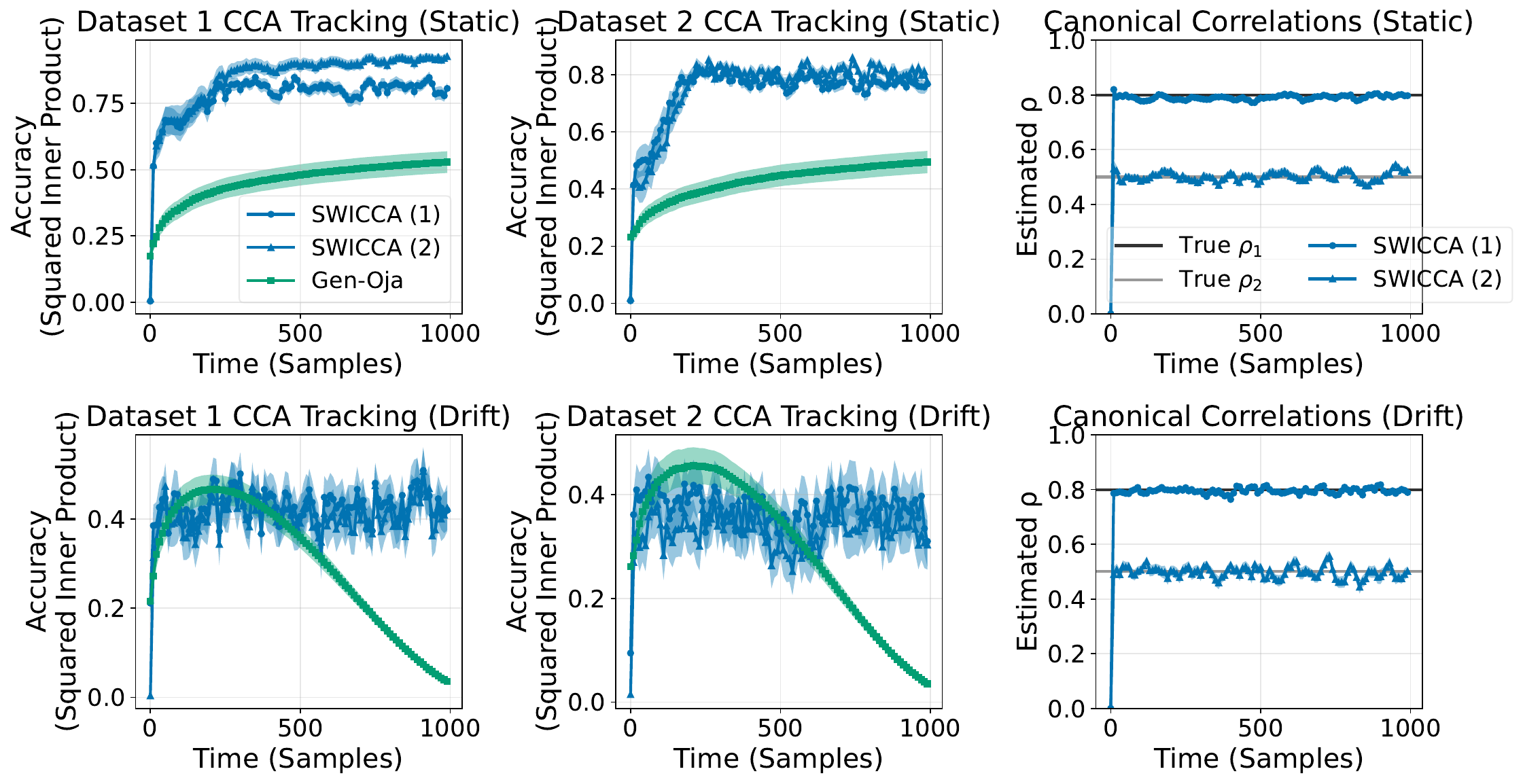}
\caption{Core tracking performance in Static and Drifting environments. 
In the static setting (top), SWICCA accurately extracts both components ($\rho_1=0.8, \rho_2=0.5$), while Gen-Oja slowly converges to the primary component. 
Under continuous drift (bottom), Gen-Oja catastrophically fails, driving its accuracy to zero as its decaying step-size prevents adaptation. 
SWICCA successfully maintains subspace alignment and accurately tracks both correlation coefficients throughout the drift.}
\label{fig:Core_Tracking}
\end{figure}

\begin{figure}[tb]
\centering
\includegraphics[width=\linewidth]{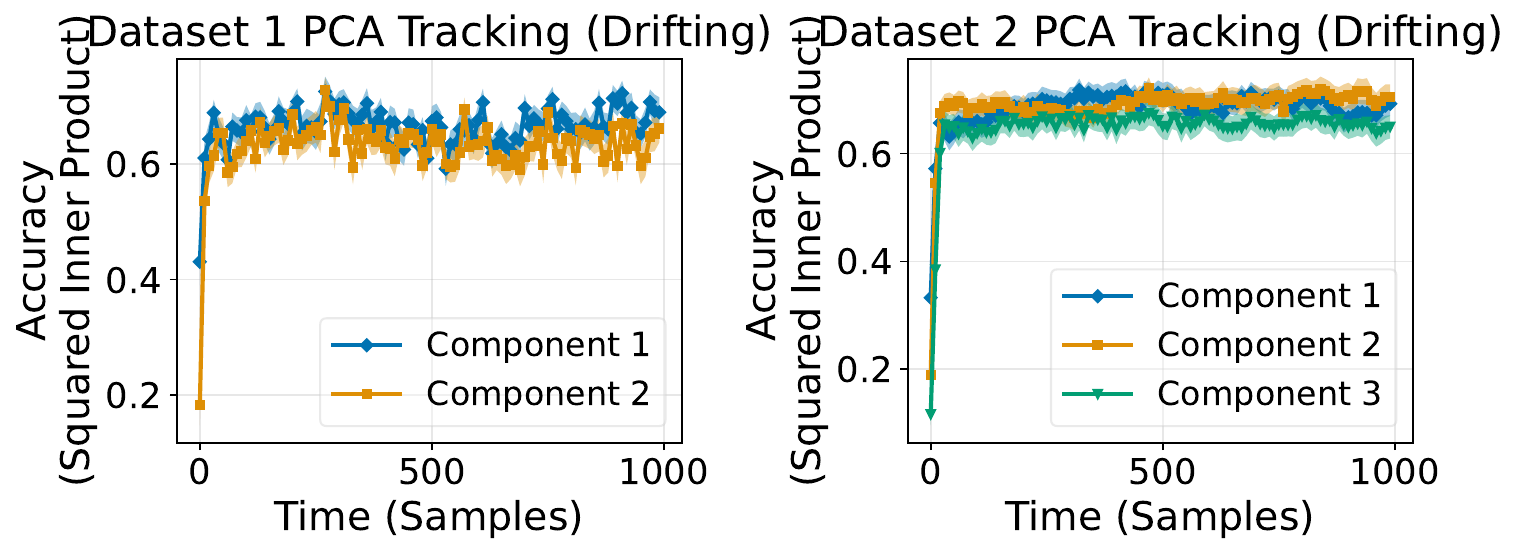}
\caption{Streaming PCA (GROUSE) performance under the continuous drift setting. 
These results represent the underlying tracking accuracy that feeds into SWICCA's correlation estimates. 
Despite the continuous rotation of the data, the components maintain a stable, high tracking accuracy.}
\label{fig:PCA_Tracking}
\end{figure}

We present the core subspace tracking results in Figure \ref{fig:Core_Tracking}. 
In the static setting, both algorithms successfully identify the primary canonical direction. 
However, SWICCA converges much faster and successfully extracts the secondary component, which Gen-Oja cannot do natively without deflation. 
In the continuous drift setting, the limitations of standard stochastic gradient methods become apparent. 
Indeed, Gen-Oja acts as a global average and completely loses the signal as the underlying subspace rotates. 
SWICCA's sliding window approach maintains agility, preserving subspace alignment and accurately tracking the correlations. 
The underlying streaming PCA performance that enables this robustness is visualized in Figure \ref{fig:PCA_Tracking}.

\begin{figure}[tb]
\centering
\includegraphics[width=\linewidth]{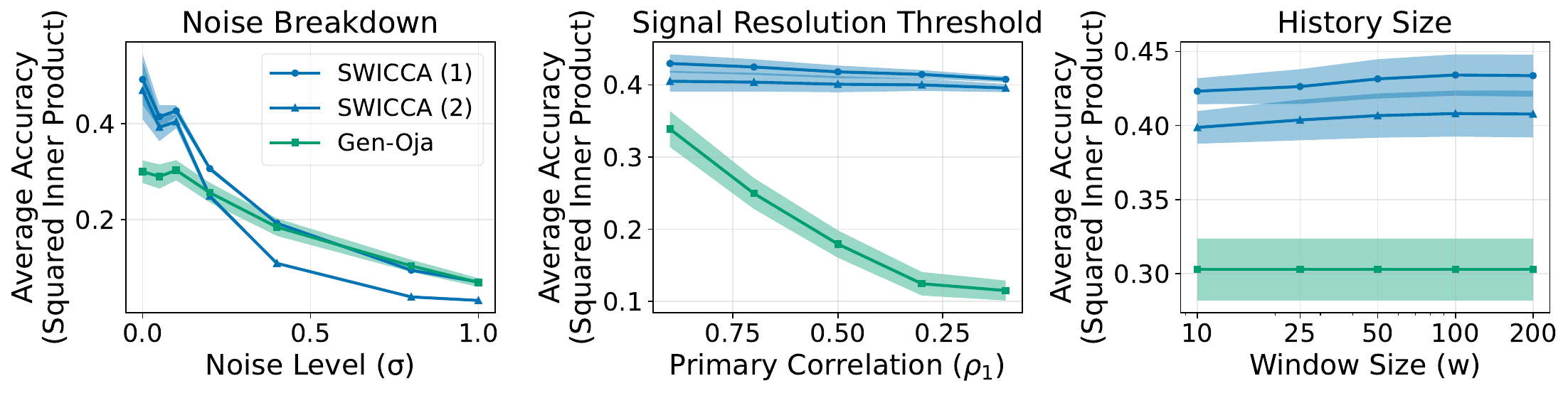}
\caption{Robustness analysis of the SWICCA algorithm. 
(Left) Tracking accuracy degrades smoothly as additive white noise ($\sigma$) increases, with SWICCA maintaining an advantage over Gen-Oja. (Center) Minimum signal resolution required for stable tracking. 
(Right) Tracking performance under continuous drift as a function of window size ($w$). 
Smaller windows respond more agilely to rapid subspace rotations, reducing lag, though a minimum $w$ is theoretically required to ensure stable, full-rank covariance estimation.}
\label{fig:Robustness}
\end{figure}

In Figure \ref{fig:Robustness}, we systematically stress-test SWICCA's robustness on drifting data. 
As we introduce and increase the additive Gaussian noise ($\sigma^2 / \sqrt{n} \, \mathcal{I}$), SWICCA's accuracy degrades smoothly rather than catastrophically, maintaining a clear performance gap over Gen-Oja at moderate noise levels. 
Furthermore, we analyze the impact of the sliding window size ($w$) under continuous drift. 
Because the underlying distributions are constantly rotating, smaller windows inherently track the immediate subspace more accurately by minimizing historical lag.
However, the window size cannot be arbitrarily small; $w$ must remain strictly greater than the latent ranks ($r_x, r_y$) to ensure the localized covariance matrices do not become rank-deficient.
Thus, utilizing a tight window (e.g., $w=25$) optimizes tracking agility while safely satisfying the algebraic requirements for stable estimation.

\begin{figure}[tb]
\centering
\includegraphics[width=\linewidth]{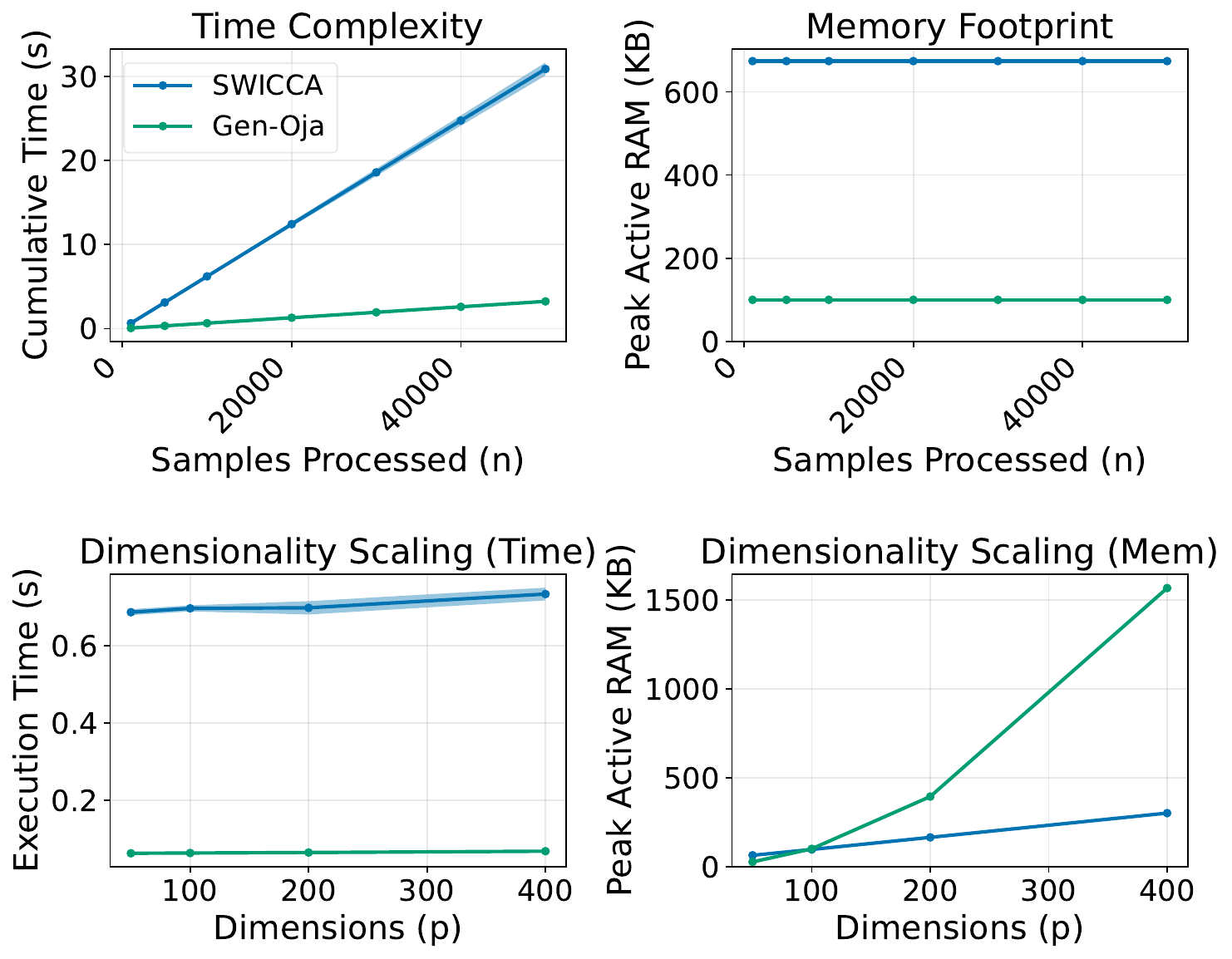}
\caption{Computational efficiency and scaling.
The first two panels demonstrate that both SWICCA and Gen-Oja execute in linear cumulative $\mathcal{O}(n)$ time and require strictly constant memory with respect to the number of samples processed.
The final two panels illustrate how execution time and peak active memory scale efficiently as the feature dimensionality ($p$) increases, validating the theoretical upper bounds established in Theorem \ref{thm:complexity}.
While Gen-Oja natively tracks a single component slightly faster and with a smaller baseline memory footprint, it fails to track non-stationary signals.
SWICCA delivers robust multi-rank adaptation while maintaining highly scalable memory and time complexities.}
\label{fig:Efficiency}
\end{figure}

Finally, we perform an empirical study of the computational efficiency of the algorithms to validate the theoretical bounds established in Theorem \ref{thm:complexity}, presented in Figure \ref{fig:Efficiency}.
We evaluate performance across varying stream lengths $n$ and feature dimensions $p$ (where $p = 2q$).
The results confirm that both methods scale linearly in cumulative time $\mathcal{O}(n)$ and strictly flatline in memory footprint with respect to the number of samples processed.
This empirical flatline directly supports our complexity analysis, as Theorem \ref{thm:complexity} indicates that SWICCA's space complexity depends only on the localized window $w$ and the dimensions, completely decoupling it from the total stream length $n$.
Furthermore, as the data dimensionality $p$ increases, the peak memory scales linearly and the per-update execution time grows very slowly, perfectly aligning with the $\mathcal{O}(p)$ theoretical upper bounds derived in our analysis.
While Gen-Oja operates with a slightly lower constant overhead due to its restriction to a single rank update, this baseline efficiency comes at the cost of catastrophic failure during data drift (as shown in Figure \ref{fig:Core_Tracking}).
In contrast, SWICCA delivers a highly adaptive, multi-rank tracking solution while remaining completely independent of the stream length $n$ in its memory requirements, requiring only fractions of a Megabyte to process the entire temporal sequence.

All simulations were run on a 2023 MacBook Pro with the Apple M2 Max processor and 32 GB of RAM; no multithreading or GPU acceleration was used. 
All algorithms were implemented in Python (version 3.14) using the \texttt{numpy} library. 
Execution times reported are wall-time, and peak active memory measurements were captured using \texttt{tracemalloc}.

\subsection{A Video Example}\label{ssec:video}

We now provide an example application of our method to a real, ultra-high-dimensional video dataset. 
We use the multi-view dataset from \cite{zheng2024DyMVHumans}, which comprises synchronized videos of a single subject performing an action from different camera angles.
In particular, we use two views (Angle 0, with the subject facing away from the camera, and Angle 25, with the subject facing the camera) from the badminton video. 
 The position and relative size of the actor in each view differ, and the actor actively moves across the frame against a static background. 
 The ideal output of CCA in this domain is to dynamically find the shared canonical subspace---in this case, isolating and tracking the correlated spatial position of the moving actor across both cameras.

This dataset is extremely high-dimensional: at $1080$p resolution, each greyscale frame contains $1088 \times 1920$ pixels, resulting in a flattened feature space of $p = 2{,}088{,}960$ dimensions per view. 
Furthermore, the video sequence is only $250$ frames long ($10$ seconds at $25$ fps). 
In the classical batch setting, solving this CCA problem is mathematically ill-posed due to the severe sample deficiency ($n \ll p$), necessitating heavily regularized alternatives. 
Additionally, the sheer scale of the data is a prohibitive computational bottleneck; naively forming the full cross-covariance matrix $C$ as in \eqref{eq:CCA_C} for a dataset of this size would require over $34$ \texttt{TB} of memory. 
Because the actor is in continuous motion, the underlying spatial correlation drifts rapidly over time, explicitly motivating a streaming, memory-efficient algorithm.

We stream the flattened video frames sequentially into the algorithms. 
To satisfy the strict constraints of online streaming without future data leakage, we do not perform global batch de-meaning. 
Instead, we maintain a streaming Exponential Moving Average (EMA) to dynamically center the data on the fly. 

We evaluate SWICCA against Gen-Oja. 
For SWICCA, we pre-specify fixed ranks of $r_1=4$ and $r_2=7$, chosen by observing a gap in the static singular value spectrum. 
We deliberately pre-specified fixed ranks to tightly control the experiment; our goal was to isolate and evaluate the numerical stability and constant memory footprint of the subspace tracking mechanism without introducing the confounding variable of fluctuating matrix sizes. 
We acknowledge that deploying this algorithm in a fully autonomous system would require wrapping the core SWICCA update with a dynamic rank-estimation method, such as a streaming adaptation of the framework proposed by \cite{nadakuditi2010fundamental, nadakuditi2011fundamental}. 
We employ a sliding window size of $w = 25$ frames.

\begin{figure}[tb]
\begin{minipage}{0.95\linewidth}
\includegraphics[width = \linewidth]{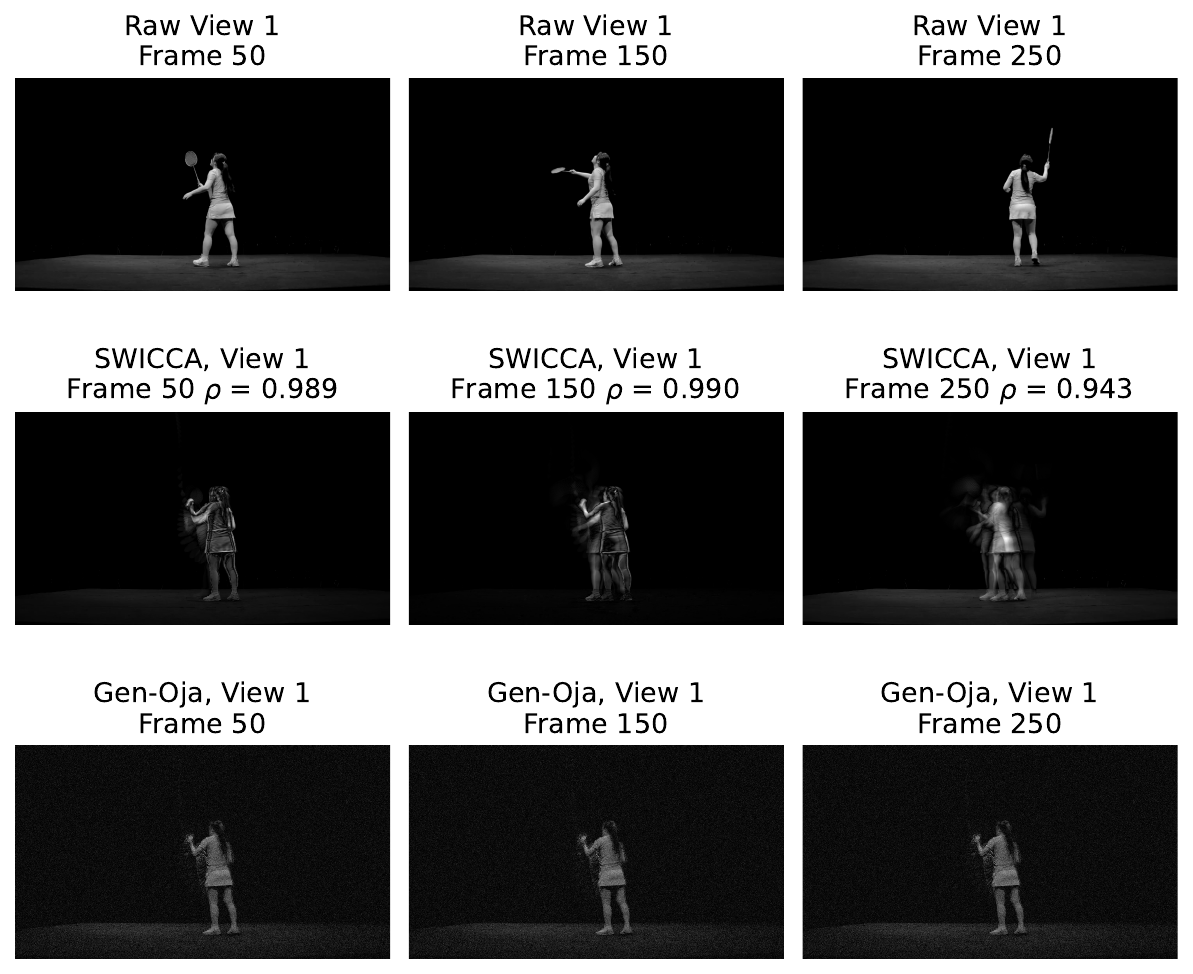}
\end{minipage}
\caption{Visualizing the canonical subspace tracking on the high-dimensional video dataset (View 1). 
The top row displays the raw video frames as the actor moves across the scene.
 The middle row displays the first spatial component extracted by SWICCA, which dynamically tracks the actor's current position with high correlation ($\rho$). 
 The bottom row displays the spatial component extracted by Gen-Oja. 
 We see that Gen-Oja averages the actor's entire historical trajectory, resulting in severe spatial smearing and an inability to localize the non-stationary signal.}
\label{fig:video_tracking}
\end{figure}

The tracking results are visualized in Figure \ref{fig:video_tracking}. 
By projecting the data onto the leading canonical components, we visualize the spatial region each algorithm identifies as highly correlated. 
SWICCA successfully tracks the movement of the actor across the frame in real-time while discarding the static background, maintaining a strong canonical correlation ($\rho > 0.94$). 
Conversely, Gen-Oja suffers catastrophic temporal smearing. 
Gen-Oja acts as a global low-pass filter, averaging the actor's historical trajectory into a single, static canonical vector that completely fails to localize the subject's current position.

We also profiled the computational efficiency of both methods. 
SWICCA successfully processed the $2$-million-dimensional streams maintaining a strictly flat memory footprint of approximately $2$ \texttt{GB} of active RAM, executing in roughly $1.7$ seconds per frame. 
While Gen-Oja executed faster ($\sim 0.1$ seconds per frame) with a lower memory profile ($\sim 200$ \texttt{MB}), it failed its primary objective of tracking the non-stationary signal. 
This confirms that SWICCA achieves highly accurate, multi-rank subspace tracking in massive streaming environments while preserving a fixed memory footprint.
While 1.7 seconds per frame on a standard CPU provides a strong baseline for algorithmic efficiency at this massive scale, achieving strict, sub-second real-time throughput for high-framerate video applications would naturally benefit from dedicated hardware acceleration or GPU integration.

\subsection{Analyzing the XRMB Dataset}\label{ssec:xrmb}

To further validate our method on diverse, non-stationary, and multi-modal real-world data, we evaluate SWICCA on the University of Wisconsin X-ray Microbeam (XRMB) Database \cite{westbury1994x}. 
This dataset provides synchronized acoustic and articulatory recordings of human speech. 
The objective of CCA in this domain is to continuously track the shared canonical subspace between the acoustic speech waveform and the physical, spatial trajectories of the speaker's vocal tract (e.g., tongue, lips, and jaw). 

We utilize a pre-processed multi-view benchmark of the XRMB data \cite{wang2014reconstruction}. 
The first view consists of 39-dimensional Mel-Frequency Cepstral Coefficients (MFCCs) extracted from the audio waveform ($p = 39$). 
The second view consists of 16-dimensional spatial coordinates representing the physical tracking pellets ($q = 16$). 
To explicitly evaluate continuous streaming tracking on a distinct phonetic sequence, we isolate and analyze a single, continuous sample---specifically, the very first complete spoken utterance in the dataset.

For this evaluation, we set the latent subspace dimensions to $r_1=5$ for the acoustic view and $r_2=4$ for the articulatory view. 
These asymmetric ranks were chosen by observing a natural gap in the static singular value spectrum of the isolated sequence. 
As in the video tracking experiment, we deliberately pre-specified these fixed ranks to tightly control the evaluation environment. 
This isolates the mathematical stability and memory footprint of SWICCA's multi-rank tracking mechanism without introducing the confounding variable of dynamically fluctuating matrix sizes. 
As in the video example, an ideal setup would include a streaming adaptation of approaches like the statistical framework proposed by \cite{nadakuditi2010fundamental, nadakuditi2011fundamental}.

This dataset presents a distinct challenge compared to the video tracking task. 
Speech production is inherently non-stationary due to coarticulation: the spatial mapping between the vocal tract and the resulting acoustics changes rapidly depending on the specific sequence of vowels and consonants. 
A static, global CCA model cannot resolve these high-frequency phonetic shifts, making this an ideal benchmark for streaming subspace tracking. 

\begin{figure}[tb]
\begin{minipage}{0.95\linewidth}
\includegraphics[width = \linewidth]{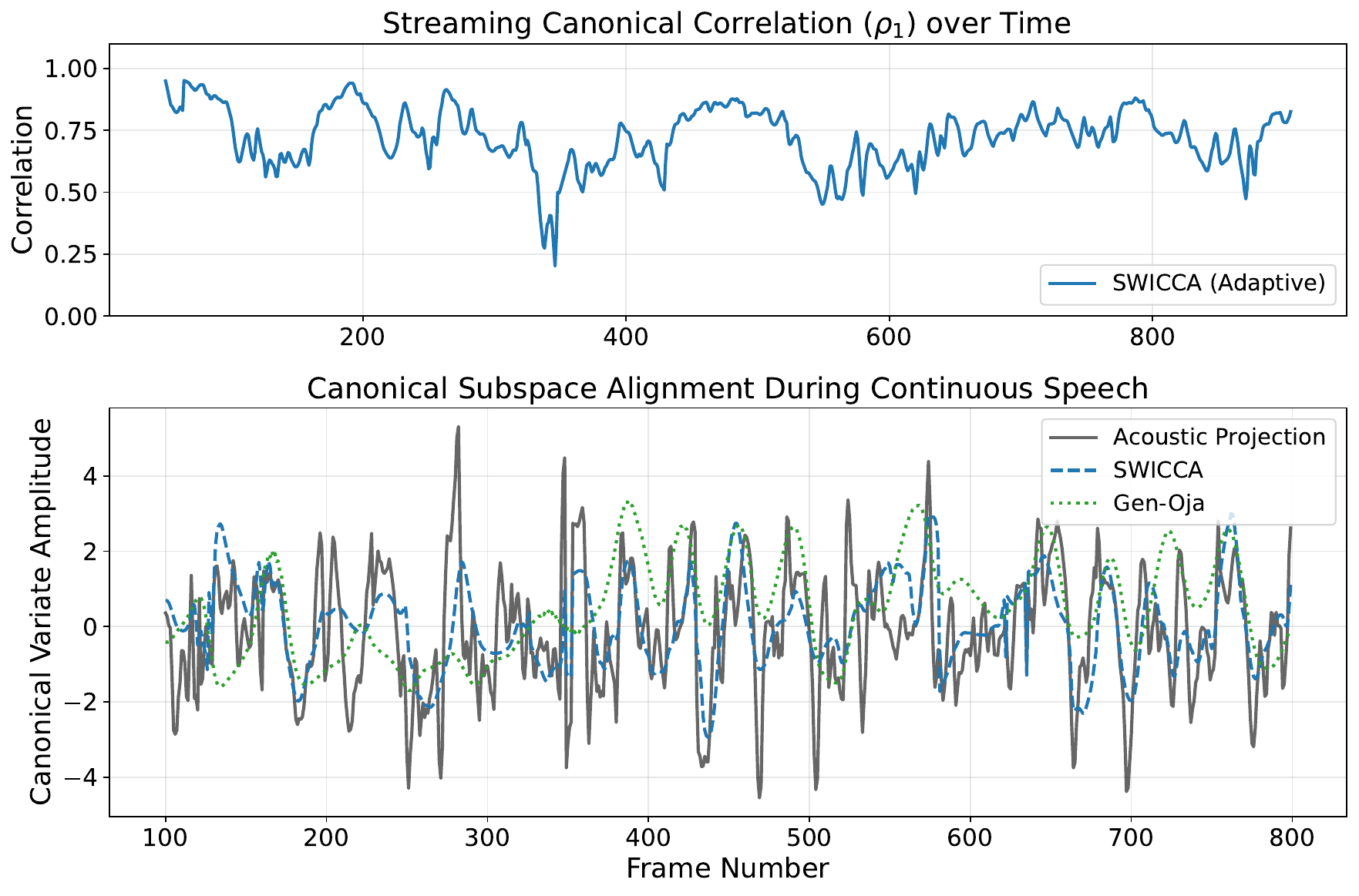}
\end{minipage}
\caption{Streaming canonical correlation analysis of the XRMB acoustic-articulatory dataset. 
(Top) SWICCA maintains a strong, dynamic correlation ($\rho_1$) over time, successfully adapting to the physical pauses and phonetic transitions in continuous speech.
 (Bottom) A visualization of the canonical subspace alignment. SWICCA's articulatory projection tightly tracks the high-frequency acoustic transients. 
 Gen-Oja fails to track the non-stationary shifts, acting as a dampened global average that becomes completely decoupled from the underlying physical speech dynamics.}
\label{fig:xrmb_tracking}
\end{figure}

The tracking results are presented in Figure \ref{fig:xrmb_tracking}. 
The top panel illustrates SWICCA's estimated primary canonical correlation ($\rho_1$) over time. SWICCA successfully maintains a strong signal lock on the speech dynamics. 
The sharp, transient dips in correlation are not algorithmic instabilities, but rather reflect the physical reality of continuous speech: during brief pauses between words or during plosive consonants (where airflow stops), the acoustic-articulatory correlation momentarily breaks down. 
SWICCA demonstrates the agility to drop its correlation during these silences and instantly re-establish a lock the moment the next syllable begins.

The bottom panel of Figure \ref{fig:xrmb_tracking} visualizes this agility directly by plotting the canonical variates (the projected signals). 
When the acoustic signal experiences sharp, rapid spikes associated with specific phonetic shifts, SWICCA's articulatory projection dynamically snaps to the correct subspace, tightly tracing the acoustic amplitude and phase.
In contrast, Gen-Oja completely fails to track the non-stationary speech. 
Because Gen-Oja functionally acts as a global average over the stream, it exhibits the behavior of a sluggish low-pass filter. 
Its articulatory projection is heavily dampened and out-of-phase, averaging conflicting phonetic mappings into a single static representation that misses the physical realities of rapid coarticulation. 
To quantify this visual discrepancy, we computed the Pearson correlation coefficient between the acoustic projection and the respective articulatory projections over the full sequence. 
SWICCA achieves a strong temporal association ($r = 0.577$), whereas Gen-Oja's globally averaged projection yields a much weaker correlation ($r = 0.295$), confirming its inability to track the localized speech dynamics.

\section{Conclusions} \label{sec:conclusions}

In this work, we introduced SWICCA, a highly adaptive, sliding-window streaming algorithm for Canonical Correlation Analysis.
We demonstrated that by operating strictly within an fixed memory footprint, our method natively extracts multi-rank canonical components without requiring computationally expensive deflation steps.
Unlike standard stochastic gradient methods, which often suffer from catastrophic temporal smearing in non-stationary environments, SWICCA maintains robust subspace alignment during continuous data drift.
Our synthetic experiments confirm that SWICCA degrades gracefully under additive noise and accurately tracks multiple underlying correlations simultaneously.
Furthermore, we validated the algorithm on two highly complex, real-world datasets.
In the ultra-high-dimensional video tracking task, SWICCA successfully isolated the spatial trajectory of a moving subject from over 2 million background pixels, maintaining stability where classical batch methods are mathematically ill-posed and computationally prohibitive.
On the XRMB speech dataset, SWICCA demonstrated the agility to resolve rapid phonetic shifts and coarticulation, tightly aligning acoustic features with their underlying physical articulatory projections to yield a temporal correlation nearly twice that of the static Gen-Oja baseline.

Alongside these empirical validations, we presented deterministic performance and error bounds for our method at a single index in time.
These bounds indicate that if the initial streaming PCA step performs well and the data are not excessively noisy, the overall performance of the SWICCA updates will remain highly accurate.
Beyond CCA, this work proposes a broader mathematical framework for adapting classical matrix decompositions to high data-rate streaming environments.
Our two-stage approach---applying streaming PCA followed by localized covariance estimation on a sliding window before performing the final decomposition---is highly extensible.
We anticipate that this sliding-window projection framework can be adapted to modernize other algorithms, including contrastive PCA (cPCA) \cite{abid2017contrastive}, cPCA++ \cite{salloum2022cpca}, and non-negative matrix factorization (NMF) \cite{wang2012nonnegative}.

While our controlled experiments utilized pre-specified ranks and tightly synchronized streams to isolate the stability of the core tracking engine, deploying SWICCA in fully autonomous environments presents several avenues for future work.
First, given assumptions about the underlying noise distributions, we imagine that our deterministic error analysis could be extended to develop sharper probabilistic performance bounds.
Second, real-world deployment will require integrating dynamic rank-estimation modules---such as those based on streaming eigenvalue thresholding---to allow the algorithm to autonomously expand and contract its latent subspace in response to shifting data regimes.
Finally, because CCA strictly depends on the temporal alignment of two datasets, we plan to investigate our algorithm's resilience to temporal jitter.
We conjecture that under relatively mild conditions on the data streams---specifically, if the autocorrelation function of each stream is non-decaying and lower bounded away from zero at finite lags---SWICCA will remain robust to small misalignments between multi-modal sensors \cite{fischer2007time, zhou2015generalized, trigeorgis2017deep, sahbi2018learning}.

\section*{Acknowledgments}
The author would like to thank Alex Foss, Uzoma Onunkwo, Cleveland Waddell, James Maissen, J. Derek Tucker, Esha Datta, Jed Duersch, and Connor Mattes for feedback on the manuscript and method. 

To ensure reproducibility and facilitate future benchmarking, it is the authors' intent to make the Python source code for SWICCA and all experimental scripts publicly available, pending the completion of formal institutional software release approvals.

This article has been authored by an employee of National Technology \& Engineering Solutions of Sandia, LLC under Contract No. DE-NA0003525 with the U.S. Department of Energy (DOE). 
The employee owns all right, title and interest in and to the article and is solely responsible for its contents. 
The United States Government retains and the publisher, by accepting the article for publication, acknowledges that the United States Government retains a non-exclusive, paid-up, irrevocable, world-wide license to publish or reproduce the published form of this article or allow others to do so, for United States Government purposes. 
The DOE will provide public access to these results of federally sponsored research in accordance with the DOE Public Access Plan \url{https://www.energy.gov/downloads/doe-public-access-plan}.

% Float barrier before appendix
\FloatBarrier
\begin{appendices}

\section{Proof of Theorem \ref{thm:complexity}: Computational Complexity}\label{app:complexity}

First, we note that $p, w \geq r_x$ and $q, w \geq r_y$.

We begin with time complexity. 
Per \cite{balzano2018streaming}, PIMC and GROUSE have a complexity of $\mathcal{O}\left(p r^2 + r^3\right)$ for a $p$-dimensional, rank $r$ fit; we use these methods as examples, and other methods can be dropped in. 
Hence, the PCA updates cost $\mathcal{O}\left(q r_y^2 + p r_x^2\right)$. 
The cost of updating the window is at most $\mathcal{O}(w + p + q)$.
Forming $X_w \widehat{V}_x$ costs $\mathcal{O}(p w r_x)$, and $Y_w \widehat{V}_y$ costs $\mathcal{O}(q w r_y)$.
The cost of normalizing each of these to form the $\widehat{U}$ and $\widehat{S}$ matrices is of order $\mathcal{O}(r_x w)$ and $\mathcal{O}(r_y w)$. 
Then, the cost of forming $\widehat{U}_x^\top \widehat{U}_y$ is $\mathcal{O}(r_x r_y w)$, and the subsequent SVD costs $\mathcal{O}(r_x r_y \min\{r_x, r_y\})$. 
Then, using the diagonal structure of $S_x$ and $S_y$, we can evaluate \eqref{eq:avoid_C} in $\mathcal{O}(p \min\{r_x, r_y\}^2 + \min\{r_x, r_y\}^2)$ and $\mathcal{O}(q \min\{r_x, r_y\}^2 + \min\{r_x, r_y\}^2)$. 
The cost of normalizing the final vectors is $\mathcal{O}(p \min\{r_x, r_y\})$ and $\mathcal{O}(q \min\{r_x, r_y\})$. 
It follows that the time complexity of each update is $\mathcal{O}(w[p r_x + q r_y])$, where we see that forming $X_w \widehat{V}_x$ and $Y_w \widehat{V}_y$ are the most expensive operations. 

We now consider the space complexity.
The storage complexity of the PCA methods is $\mathcal{O}\left(p r_x + q r_y\right)$. 
The cost of the window is $\mathcal{O}(w [ p + q ])$. 
Storing the $\widehat{U}$ and $\widehat{S}$ costs $\mathcal{O}(w r_x + r_x^2)$ and $\mathcal{O}(w r_y + r_y^2)$.
Storing $\widehat{U}_x^\top \widehat{U}_y$ costs $\mathcal{O}(r_x r_y)$ and the subsequent SVD costs $\mathcal{O}(r_x^2 + r_y^2 +  \min\{r_x, r_y\}^2)$.
The final output vectors have costs of $\mathcal{O}(p  \min\{r_x, r_y\})$ and $\mathcal{O}(q  \min\{r_x, r_y\})$.
It follows that the total space required scales as $\mathcal{O}(w [p + q])$, where we see that storing the window of data is the dominant factor. 

Note that if we were to store a window of loadings instead, the storage of the window would drop of $\mathcal{O}(w r_x + w r_y)$ and the overall space complexity would change to $\mathcal{O}(\max\{p, w\} r_x + \max\{q, w\} r_y)$. 
The time complexity of forming $X_w \widehat{V}_x$ and $Y_w \widehat{V}_y$ also drops to $\mathcal{O}(p r_x)$ and $\mathcal{O}(q r_y)$, as we now only update a single row at a time; the rest of the operations are identical, but the time complexity per iteration drops to $\mathcal{O}(p r_x^2 + q r_y^2 + w [r_x + r_y] + r_x r_y \min\{r_x, r_y\})$.
If the ranks $r_x$ and $r_y$ are small, we may drop the last term.

\section{Proof of Theorem \ref{thm:error}: Error Analysis}\label{app:error}

In what follows, we prove Theorem \ref{thm:error}. 
At a high level, the proof steps through the SWICCA algorithm and decomposes and propagates the error at each step.

\subsection{Error matrices}

We see that an estimate of $U_x S_x$ is obtained by
\begin{subequations}
\begin{equation}
U_x S_x \widehat{V}_x = U_x S_x + X \Delta_x,
\end{equation}
and similarly for $U_y S_y$. We may now obtain estimates for $U_x$ by normalizing the columns of $U_x S_x$. We may then write
\begin{equation}
\widehat{U_x} = U_x + \Delta_{U_x} \in \RR^{w \times r_x},
\end{equation}
where $\Delta_{U_x}$ is a function of $X \Delta_x$ (and similarly for $U_y$). To estimate $S_x$, we use the norms of the columns of our estimate of $U_x S_x$, and we may write
\begin{equation}
\widehat{S_x} = S_x + \Delta_{S_x} \in \RR^{r_x \times r_x},
\end{equation}
and similarly for $S_y$; note that the estimates and the errors for $S_x$ are diagonal, and that the error is once again a function of $X \Delta_x$.

The next step of the algorithm forms the CCA matrix $C$. We may write
\begin{equation}
\begin{split}
\widehat{C} &= (V_x + \Delta_x) (U_x + \Delta_{U_x})^\top (U_y + \Delta_{U_y}) (V_y + \Delta_y)^\top \\
&= \left(V_x U_x^\top + \Delta_x U_x^\top + V_x \Delta_{U_x}^\top + \Delta_x \Delta_{U_x}^\top\right) \\
&\qquad \left(U_y V_y^\top + \Delta_{U_y} V_y^\top + U_y \Delta_y^\top + \Delta_{U_y} \Delta_y^\top\right) \\
&= V_x U_x^\top U_y V_y^\top \\
&\quad + V_x U_x^\top \left(\Delta_{U_y} V_y^\top + U_y \Delta_y^\top + \Delta_{U_y} \Delta_y^\top\right) \\
&\quad + \left(\Delta_x U_x^\top + V_x \Delta_{U_x}^\top + \Delta_x \Delta_{U_x}^\top\right) U_y V_y^\top \\
&\quad + \left(\Delta_x U_x^\top + V_x \Delta_{U_x}^\top + \Delta_x \Delta_{U_x}^\top\right) \\
&\qquad \left(\Delta_{U_y} V_y^\top + U_y \Delta_y^\top + \Delta_{U_y} \Delta_y^\top\right) \\
&= V_x U_x^\top U_y V_y^\top + \Delta_C,
\end{split}
\end{equation}
where we have implicitly defined $\Delta_C \in \RR^{p \times q}$. 

Before looking at the SVD of $\widehat{C}$, we look at the other matrix that is estimated. We may write
\begin{equation}
\begin{split}
\widehat{\Sigma_x^{-1/2}} &= (V_x + \Delta_x) (S_x + \Delta_{S_x})^{-1} (V_x + \Delta_x)^\top \\
&= (V_x + \Delta_x) (S_x^{-1} + \Delta_{S_x^{-1}}) (V_x + \Delta_x)^\top \\
&= V_x S_x^{-1} V_x^\top + V_x S_x^{-1} \Delta_x^\top \\
&\quad + V_x \Delta_{S_x^{-1}}\Delta_x^\top + V_x \Delta_{S_x^{-1}}V_x^\top  + \Delta_x S_x^{-1} \Delta_x^\top \\
&\quad + \Delta_x \Delta_{S_x^{-1}}\Delta_x^\top + \Delta_x \Delta_{S_x^{-1}} V_x^\top + \Delta_x S_x^{-1} V_x^\top \\
&= V_x S_x^{-1} V_x^\top + \Delta_{\Sigma_x^{-1/2}},
\end{split}
\end{equation}
where we have implicitly defined $\Delta_{\Sigma_x^{-1/2}} \in \RR^{p \times p}$. A similar expression holds for $\widehat{\Sigma_y^{-1/2}}$.

\end{subequations}

\subsection{Bounding the error matrices}\label{ssec:bound_err_mat}

\begin{subequations}

We will now bound the sizes of the various error matrices derived above. 
We will present results and derivations for $X$ and note that analogous results will hold for $Y$. 

We may write a column of $U_x S_x \widehat{V}_x$ as 
\begin{equation}
\sigma_{x, i} \uu_{x, i} + (X \Delta_x)_i,
\end{equation}
where $ (X \Delta_x)_i$ denotes the $i^{th}$ column of $(X \Delta_x)$. 
The norm of this vector is the estimated singular value $\widehat{\sigma}_{x, i}$, and is bounded by 
\begin{equation}
\left|\sigma_{x, i} - \left\| (X \Delta_x)_i \right\|_2\right| \leq \widehat{\sigma}_{x, i} \leq \sigma_{x, i} + \left\| (X \Delta_x)_i \right\|_2.
\end{equation}
It follows that
\begin{equation}\label{eq:sigma_err}
\left|\sigma_{x, i} - \widehat{\sigma}_{x, i}\right| \leq \min\left\{\left\| (X \Delta_x)_i \right\|_2, \sigma_{x, i}\right\}.
\end{equation}

Next, the normalized vector above is our estimate of $\uu_{x, i}$, and we may write
\begin{equation}
\left\| \uu_{x, i} - \widehat{\uu}_{x, i}\right\|_2 = \left\| \left(1 - \frac{\sigma_{x, i}}{\widehat{\sigma}_{x, i}}\right) \uu_{x, i} + \frac{1}{\widehat{\sigma}_{x, i}} (X \Delta_x)_i\right\|_2.
\end{equation}
We may bound
\begin{equation}
\begin{split}
\left|1 - \frac{\sigma_{x, i}}{\widehat{\sigma}_{x, i}}\right| &\leq \frac{1}{\widehat{\sigma}_{x, i}} \min\left\{\left\| (X \Delta_x)_i \right\|_2, \sigma_{x, i}\right\} \\
&\leq \frac{\min\left\{\left\| (X \Delta_x)_i \right\|_2, \sigma_{x, i}\right\}}{\left|\sigma_{x, i} - \left\| (X \Delta_x)_i \right\|_2\right|}
\end{split}
\end{equation}
so that 
\begin{equation}\label{eq:u_err}
\left\| \uu_{x, i} - \widehat{\uu}_{x, i}\right\|_2 \leq \frac{\left\| (X \Delta_x)_i \right\|_2 + \min\left\{\left\| (X \Delta_x)_i \right\|_2, \sigma_{x, i}\right\}}{\left|\sigma_{x, i} - \left\| (X \Delta_x)_i \right\|_2\right|}.
\end{equation}

We next look at the error in the reciprocals of the singular values. 
We may write
\begin{equation}
\left|\frac{1}{\sigma_{x, i}} - \frac{1}{\widehat{\sigma}_{x, i}}\right| = \frac{\left|\widehat{\sigma}_{x, i} - \sigma_{x, i}\right|}{\sigma_{x, i} \widehat{\sigma}_{x, i}}.
\end{equation}
Using the derivations above, we have that 
\begin{equation}\label{eq:inv_sigma_err}
\left|\frac{1}{\sigma_{x, i}} - \frac{1}{\widehat{\sigma}_{x, i}}\right|  \leq \frac{ \min\left\{\left\| (X \Delta_x)_i \right\|_2, \sigma_{x, i}\right\} }{\sigma_{x, i} \left|\sigma_{x, i} - \left\| (X \Delta_x)_i \right\|_2\right|}.
\end{equation}

\end{subequations}

\subsubsection{Intermediate Takeaways}

\begin{subequations}

While our error analysis is not yet complete, we are able make the following conclusions. 
By summing the terms in (\ref{eq:sigma_err}), (\ref{eq:u_err}), and (\ref{eq:inv_sigma_err}), we may bound $\left\|\Delta_{S_x}\right\|_F$, $\left\|\Delta_{U_x}\right\|_F$, and $\left\|\Delta_{S_x^{-1}}\right\|_F$, respectively. In particular, we may write the following rather loose bounds:
\begin{equation}
\left\|\Delta_{S_x}\right\|_F \leq  \min\left\{\left\|X \Delta_x\right\|_F, r_x \sigma_{x, 1}\right\},
\end{equation}
\begin{equation}
\left\|\Delta_{U_x}\right\|_F \leq \frac{ \left\|X \Delta_x\right\|_F + \min\left\{\left\|X \Delta_x\right\|_F, r_x \sigma_{x, 1}\right\} }{\min_{1 \leq i \leq r_x} \left|\sigma_{x, i} - \left\| (X \Delta_x)_i \right\|_2\right| },
\end{equation}
\begin{equation}
\left\|\Delta_{S_x^{-1}}\right\|_F \leq \frac{ \min\left\{\left\|X \Delta_x\right\|_F, r_x \sigma_{x, 1}\right\} }{c_{\sigma} \min_{1 \leq i \leq r_x} \left|\sigma_{x, i} - \left\| (X \Delta_x)_i \right\|_2\right| }.
\end{equation}
While the presence of 
\begin{equation}
\left|\sigma_{x, i} - \left\| (X \Delta_x)_i\right\| \right|
\end{equation}
in denominator of these bounds might appear problematic, we quickly see that since the non-zero singular values are bounded  by 
\begin{equation}
C_{\sigma} \geq \sigma_{x, 1} \geq \sigma_{x, i} > c_{\sigma} > 0
\end{equation}
(for $1 \leq i \leq r_x$) and the rank $r_x$ is fixed, as long as $\left\|X \Delta_x\right\|_F \rightarrow 0$, so will each of $\left\|\Delta_{S_x}\right\|_F$, $\left\|\Delta_{U_x}\right\|_F$, and $\left\|\Delta_{S_x^{-1}}\right\|_F$.

\end{subequations}

\subsubsection{Bounding \texorpdfstring{$\left\|\Delta_C\right\|_F$}{the error in C}}

\begin{subequations}

Using the unitary invariance of the Frobenius norm (or, the Cauchy Schwarz inequality) and the triangle inequality, we see that 
\begin{equation}
\begin{split}
\left\|\Delta_C\right\|_F &\leq \left(\left\|\Delta_{U_x}\right\|_F + \left\|\Delta_{x}\right\|_F + \left\|\Delta_{U_x}\right\|_F \left\|\Delta_{x}\right\|_F\right)  \\
&\quad + \left(\left\|\Delta_{U_y}\right\|_F + \left\|\Delta_{y}\right\|_F + \left\|\Delta_{U_y}\right\|_F \left\|\Delta_{y}\right\|_F\right)\\
&\quad + \left(\left\|\Delta_{U_y}\right\|_F + \left\|\Delta_{y}\right\|_F + \left\|\Delta_{U_y}\right\|_F \left\|\Delta_{y}\right\|_F\right)\\
&\qquad \left(\left\|\Delta_{U_x}\right\|_F + \left\|\Delta_{x}\right\|_F + \left\|\Delta_{U_x}\right\|_F \left\|\Delta_{x}\right\|_F\right).
\end{split}
\end{equation}
There are two terms and their product in the above expression. 
We may bound the first term by
\begin{equation}
\begin{split}
&\left(\left\|\Delta_{U_x}\right\|_F + \left\|\Delta_{x}\right\|_F + \left\|\Delta_{U_x}\right\|_F \left\|\Delta_{x}\right\|_F\right) \\
&\quad \leq \frac{ 2 \left\|X \Delta_x\right\|_F \left(1 + \left\|\Delta_x\right\|_F\right)}{\min_{1 \leq i \leq r_x} \left|\sigma_{x, i} - \left\| (X \Delta_x)_i \right\|_2\right| } + \left\|\Delta_{x}\right\|_F,
\end{split}
\end{equation}
 and similarly for the second term with $y$ replacing $x$. If we define
\begin{equation}
\begin{split}
\eta_{C, xy} = \max \Bigg\{ &\frac{ 2 \left\|X \Delta_x\right\|_F \left(1 + \left\|\Delta_x\right\|_F\right)}{\min_{1 \leq i \leq r_x} \left|\sigma_{x, i} - \left\| (X \Delta_x)_i \right\|_2\right| } + \left\|\Delta_{x}\right\|_F, \\
&\frac{ 2 \left\|Y \Delta_y\right\|_F \left(1 + \left\|\Delta_y\right\|_F\right)}{\min_{1 \leq i \leq r_y} \left|\sigma_{y, i} - \left\| (Y \Delta_y)_i \right\|_2\right| } + \left\|\Delta_{y}\right\|_F \Bigg\}
\end{split}
\end{equation}
 we may then bound $\left\|\Delta_C\right\|_F$ by
 \begin{equation}
 \left\|\Delta_C\right\|_F \leq 2 \eta_{C, xy} + \eta_{C, xy}^2.
 \end{equation}
Hence, define
\begin{equation}\label{eq:eta_C}
\begin{split}
\eta_{C} = \max \Big\{ & \left\|X \Delta_x\right\|_F, \left\|Y \Delta_x\right\|_F, \left\|\Delta_x\right\|_F, \left\|\Delta_y\right\|_F, \\
& \left\|X \Delta_x\right\|_F \left\|\Delta_x\right\|_F, \left\|Y \Delta_y\right\|_F \left\|\Delta_y\right\|_F \Big\},
\end{split}
\end{equation}
so that if $\max\left\{\eta_C, \eta_C^2\right\} \rightarrow 0$, we have that $\left\|\Delta_C\right\|_F \rightarrow 0$. 
 
 \end{subequations}
 
 \subsubsection{Bounding \texorpdfstring{$\left\|\Delta_{\Sigma_x^{-1/2}}\right\|_F$}{the error in the inverse-square-root covariance}}
 
 \begin{subequations}
 
 We may perform a similar analysis as in the previous section, and find that 
\begin{equation}
\begin{split}
\Delta_{\Sigma_x^{-1/2}} &\leq 2 \left\|S_x^{-1}\right\|_F \left\|\Delta_x\right\|_F + 2 \left\|\Delta_{S_x^{-1}} \right\|_F \left\|\Delta_x\right\|_F \\ 
&\quad + \left\|\Delta_{S_x^{-1}}\right\|_F + \left\|S_x^{-1}\right\|_F \left\|\Delta_x\right\|_F^2 \\
&\quad + \left\|\Delta_{S_x^{-1}} \right\|_F \left\|\Delta_x\right\|_F^2.
\end{split}
\end{equation}
Similarly, defining
\begin{equation}\label{eq:eta_sigma}
\begin{split}
\eta_{\Sigma_x} = \max \Big\{ & \left\|X \Delta_x\right\|_F, \left\|X \Delta_x\right\|_F \left\|\Delta_x\right\|_F, \left\|\Delta_x\right\|_F, \\
& \left\|X \Delta_x\right\|_F \left\|\Delta_x\right\|_F^2, \left\|\Delta_x\right\|_F^2 \Big\},
\end{split}
\end{equation}
yields that if $\eta_{\Sigma_x} \rightarrow 0$, then 
\begin{equation}
\left\|\Delta_{\Sigma_x^{-1/2}}\right\|_F \rightarrow 0.
\end{equation}
A similar condition holds with $y$ replacing $x$. 

\end{subequations}

\subsubsection{Bounding the error in the SVD of \texorpdfstring{$\widehat{C}$}{the estimated matrix C}}

\begin{subequations}

The penultimate step in the SWICCA algorithm is to take the SVD of $\widehat{C}$. 
Invoking the results from \cite[Theorem~3]{yu2015useful}, we may bound the errors in the estimates of $W$ and $H$ as follows: there exist an orthogonal matrices $O_W$ and $O_H$ such that 
\begin{equation}
\begin{split}
\left\|\Delta_W\right\|_F &= \left\|\widehat{W} O_W - W\right\|_F \\
&\leq \frac{2^{3/2} \left(2 \sigma_{C, 1} + \left\|\Delta_{C}\right\|_2\right) \left\|\Delta_{C}\right\|_F}{\sigma_{C, r_C}^2},
\end{split}
\end{equation}

\begin{equation}
\begin{split}
\left\|\Delta_H\right\|_F &= \left\|\widehat{H} O_H - H\right\|_F \\
&\leq \frac{2^{3/2} \left(2 \sigma_{C, 1} + \left\|\Delta_{C}\right\|_2\right) \left\|\Delta_{C}\right\|_F}{\sigma_{C, r_C}^2},
\end{split}
\end{equation}
where $\sigma_{C, i}$ denotes the $i^{th}$ singular value of the matrix $C$. 

Note that this result depends on the weakest correlation, encoded by the smallest singular value of $C$. It follows that when 
\begin{equation}
\eta_{WH} = \max\left\{\eta_C, \eta_C^2, \eta_C^3\right\} \rightarrow 0,
\end{equation}
we have that $\left\|\Delta_W\right\|_F, \left\|\Delta_H\right\|_F \rightarrow 0$, where $\eta_C$ was defined in (\ref{eq:eta_C}). 

\end{subequations}

\subsubsection{Final statement of error bounds}

\begin{subequations}

We may now package together the error bounds from the SVD of $\widehat{C}$ and the estimates of $\Sigma_x^{-1/2}$ and $\Sigma_y^{-1/2}$. 
We may repeat the analysis in section \ref{ssec:bound_err_mat}, where we bounded the deviation $\left\|\uu_{x, i} - \widehat{\uu}_{x, i}\right\|_2$, and replace $\uu_{x, i}$ with $\ff_i$, $\sigma_{x, i}$ with $\left\|\Sigma_{x}^{-1/2} \ww_i\right\|_2$, and $X \Delta_x$ with 
\begin{equation}
\Delta_{\Sigma_x^{-1/2}} W + \Sigma_x^{-1/2} \Delta_{W} + \Delta_{\Sigma_x^{-1/2}} \Delta_W,
\end{equation}
and similarly for $H$. 

It follows that if
\begin{equation}
\max\left\{\eta_{WH}, \eta_{\Sigma_x}, \eta_{\Sigma_y}, \eta_{WH} \eta_{\Sigma_x}, \eta_{WH} \eta_{\Sigma_y}\right\} \rightarrow 0,
\end{equation}
then 
\begin{equation}
\left\|O_W \widehat{F} - F\right\|_F \rightarrow 0,
\end{equation}
\begin{equation}
\left\|O_H \widehat{G} - G\right\|_F \rightarrow 0,
\end{equation}
as desired, where we have defined $\eta_{\Sigma_x}$ in (\ref{eq:eta_sigma}). 
Moreover, the singular values of $\widehat{C}$ will similarly be close to those of $C$ under the same conditions, and we note that these singular values are estimators of the absolute values of the correlations \cite{asendorf2017improved}. 

Finally, we end by noting that all of the above conclusions hold if
\begin{equation}
\max\left\{\left\|\Delta_x\right\|_F, \left\|\Delta_y\right\|_F, \left\|X \Delta_x\right\|_F, \left\|Y \Delta_y\right\|_F\right\} \rightarrow 0.
\end{equation}
Moreover, if we assume that the smallest singular value of $C$ ($\sigma_{r_C}$) is lower bounded by some absolute constant $c_{\rho} > 0$, we may allow the correlation values to drift. 

In conclusion, we have shown that if the streaming PCA algorithm yields accurate or consistent estimates of the principal components and if the noise level in the data is not too high relative to the error in the PCA estimates, the SWICCA will also produce accurate estimates. 

\end{subequations}

\end{appendices}

\bibliographystyle{IEEEtran} 
\bibliography{SWICCA}

\vspace{-1.15cm}
\begin{IEEEbiography}[{\includegraphics[width=1in, height=1.25in, clip, keepaspectratio]{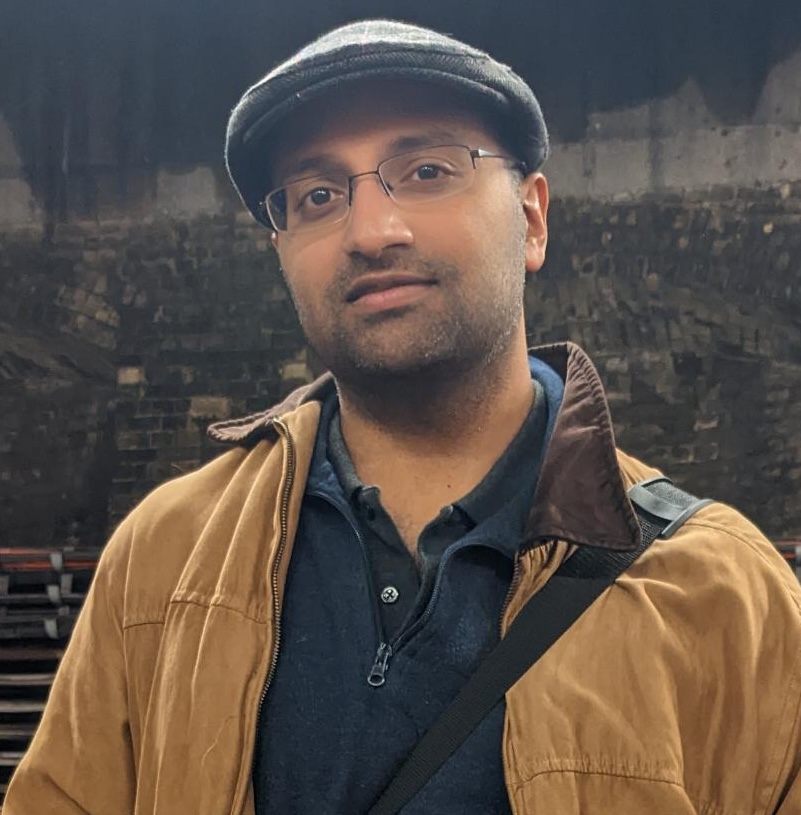}}]{Arvind Prasadan}
received the B.S. degree in Electrical Engineering and Computer Science and in Mathematics from the University of Pittsburgh in 2014, and the Ph.D. degree in Electrical Engineering: Systems from the University of Michigan, Ann Arbor in 2020. 
Since 2020, he has been at Sandia National Laboratories in Livermore, CA. 
His research interests include audio signal processing; real-time, online, and streaming machine learning; applications of random matrix theory; tensor decompositions; and machine learning security. 
\end{IEEEbiography}

\vfill

\end{document}